\documentclass[letterpaper]{article} 
\usepackage[submission]{aaai25}  
\usepackage{times}  
\usepackage{helvet}  
\usepackage{courier}  
\usepackage[hyphens]{url}  
\usepackage{graphicx} 
\usepackage{natbib}  
\usepackage{amsfonts}
\usepackage{xcolor}
\usepackage{caption} 
\usepackage{algorithm}
\usepackage{algorithmic}

\usepackage{newfloat}
\usepackage{listings}
\usepackage{amsmath,graphicx,tabularx,multirow,booktabs,latexsym}
\DeclareCaptionStyle{ruled}{labelfont=normalfont,labelsep=colon,strut=off} 
\floatstyle{ruled}
\newfloat{listing}{tb}{lst}{}
\floatname{listing}{Listing}
\title{TRRG: Towards Truthful Radiology Report Generation With
 Cross-modal Disease Clue Enhanced Large Language Models}
\author {
    Yuhao Wang\textsuperscript{\rm 1},
    Chao Hao\textsuperscript{\rm 1},
    Yawen Cui\textsuperscript{\rm 2},
    Xinqi Su\textsuperscript{\rm 1},
    Weicheng Xie\textsuperscript{\rm 3},
    Tao Tan\textsuperscript{\rm 4},
    Zitong Yu\textsuperscript{\rm 1 }\thanks{Corresponding Author}
}
\affiliations {
    \textsuperscript{\rm 1}Great Bay University \quad
    \textsuperscript{\rm 2}The Hong Kong Polytechnic University \quad
    \textsuperscript{\rm 3}Shenzhen University \quad
    \textsuperscript{\rm 4}Macao Polytechnic  University
    
}

\usepackage{bibentry}

\begin{document}

\maketitle

\begin{abstract}
The vision-language modeling capability of multi-modal large language models has attracted wide attention from the community. However, in medical domain, radiology report generation using vision-language models still faces significant challenges due to the imbalanced data distribution caused by numerous negated descriptions in radiology reports and issues such as rough alignment between radiology reports and radiography. In this paper, we propose a truthful radiology report generation framework, namely TRRG, based on stage-wise training for cross-modal disease clue injection into large language models. In pre-training stage, During the pre-training phase, contrastive learning is employed to enhance the visual encoder's ability to perceive fine-grained disease details. In fine-tuning stage, the clue injection module we proposed significantly enhances the disease-oriented perception capability of the large language model by effectively incorporating the robust zero-shot disease perception. Finally, through the cross-modal clue interaction module, our model effectively achieves the multi-granular interaction of visual embeddings and an arbitrary number of disease clue embeddings. This significantly enhances the report generation capability and clinical effectiveness of multi-modal large language models in the field of radiology reportgeneration. Experimental results demonstrate that our proposed pre-training and fine-tuning framework achieves state-of-the-art performance in radiology report generation on datasets such as IU-Xray and MIMIC-CXR. Further analysis indicates that our proposed method can effectively enhance the model’s ability to perceive diseases and improve its clinical effectiveness.
\end{abstract}

%

\section{Introduction}
Radiology report generation aims to automatically generate radiology reports for medical images, such as X-rays, MRI, and CT scans, etc. Radiology report generation can effectively alleviate the workload of clinical practitioners, minimize misdiagnosis and missed diagnoses resulting from human judgment errors, and expedite clinical workflows. A radiology report typically consists of a paragraph with multiple sentences. Some sentences describe normal organ appearances in radiography, while others delineate the findings of diseases reflected in the images, including the location of lesions, types of diseases, severity, etc. With the development of visual-language models, encoder-decoder-based text generation models have become the most prevalent architecture. However, unlike natural domain image captioning\cite{tang2021clip4caption,2017Knowing,rennie2017selfcritical,pan2020xlinear,johnson2015densecap}, radiology reports contain a significant amount of normal descriptions. Abnormal descriptions related to diseases are crucial for radiology reports. This leads to sparse supervision signals regarding diseases. Many pure encoder-decoder-based models\cite{li2019knowledgedriven,chen2020generating,Wang_2021_CVPR,2021Knowledge,wang2023self,2020icdmAutomatic,Knowledge-Driven,CVPR201_PPKD} for image-text coarse-grained alignment lack fine-grained perception of diseases. Consequently, these models fail to effectively focus on the correct image regions, resulting in insufficient clinical effectiveness of the generated radiology reports. Multi-modal large language models\cite{li2023blip2,alayrac2022flamingo,dai2023instructblip,chen2024lion}, as powerful infrastructure adaptable to various downstream tasks, perform well in tasks such as image captioning and Visual Question Answering (VQA). Through techniques such as low-parameter fine-tuning, these models can efficiently enhance multimodal large language models for radiology report generation.

To further enhance the clinical effectiveness of radiology report generation (RRG), we propose a stage-wise, cross-modal clue-injection approach based on large language models. Firstly, we perform pretraining using image-text contrastive learning \cite{CLIP} through sampling sentences from radiology reports to enhance the vision encoder's ability to perceive diseases effectively. Moreover, by harnessing the powerful zero-shot inference capability of the vision encoder, our model can effectively acquire clues about various diseases. In the fine-tuning stage, we conduct clue injection through visual disease tokens and several disease clues embeddings. Subsequently, we propose a cross-modal clue interaction module that effectively integrates visual embeddings and disease clue embeddings to enhance the disease perception capability of the large language model. Finally, we propose a disease-aware consistency loss to assist large language models in training for radiology report generation tasks. The disease-aware consistency loss effectively enhances textual semantic supervision signals, promoting the model to acquire disease-awareness capability while generating radiology reports. Our model achieves optimal performance in generating radiology reports. We conduct experiments on the IU-Xray \cite{2015iu-xray} and MIMIC-CXR \cite{2019MIMIC} datasets. Compared with previous studies, our method achieves state-of-the-art results, significantly improving both the quality of language generation and clinical effectiveness. Further ablation experiments validate the effectiveness of our proposed disease clue injection module, cross-modal clue interaction module, and disease-aware consistency loss. Qualitative results offer intuitive interpretations of the generated radiology reports.

\begin{itemize}

\item We propose the TRRG for truthful radiology report generation using an disease clue injection enhanced large language model. TRRG alleviates the problem of coarse-grained alignment between radiography and report, enabling the model acquire fine-grained disease-aware perception.

\item We introduce a cross-modal disease clue interaction module, which effectively integrates visual embeddings and disease clue embeddings to guide large language models in producing higher-quality radiology reports.

\item Experimental results demonstrate that our proposed method outperforms previous approaches in terms of both language generation quality and clinical effectiveness on two datasets, IU-Xray \cite{2015iu-xray} and MIMIC-CXR \cite{2019MIMIC}. 

\end{itemize}

\section{Related work}
\subsubsection{Radiology Report Generation}
Currently, most research on radiology report generation adopts an encoder-decoder architecture. Some studies employ CNN-RNN architecture \cite{2018Multimodal, li2019knowledgedriven, 2020When, 2019muiti-view}. Through convolution neural network encoders, these models encode images into vectors, which are then decoded token by token using recurrent neural networks. Some studies utilize a hierarchical generation process \cite{2020icdmAutomatic}, in which the model initially generates relevant topics as keywords. Subsequently, it employs LSTM architecture to generate sentences for each keyword. With the advancement of transformers \cite{vaswani2017attentionisallyouneed}, which benefit from parallel training, are suitable for modeling long sequences, and have the capacity to integrate various modalities of visual and language data, radiology report generation is gradually transitioning towards architectures based on transformers \cite{CVPR201_PPKD,2019MultimodalTransformer,chen2020generating,chen2022cross,cornia2020meshedmemory,wang2023metransformer}. Some studies utilize memory networks \cite{chen2022cross, chen2020generating, 2020Meshed} to help the model learn more effective patterns of medical knowledge. For example, use a memory matrix to simulate memory information and continuously update this matrix during training to generate higher-quality radiology reports. Others adopt prototype learning\cite{wang2022cross}, storing prototype vectors and updating them through cross-modal queries and responses to drive the model to store pattern information in memory. Furthermore, some research enhances the quality of generated radiology reports by incorporating external knowledge embeddings \cite{Knowledge-Driven,2021Knowledge}. This involves constructing knowledge graphs of anatomical structures and disease-related relationships as auxiliary features for embedding, which are then combined with the model for the final text generation. Historical radiography-report pairs \cite{KnowledgeMatters} are also utilized as domain knowledge, mimicking the process of doctors generating textual reports, thereby improving the performance of the respective models during decoding. With the advancement of large language models. However, all these coarse-grained alignment methods lead to limitations in the clinical effectiveness of the generated radiology reports.

\subsubsection{Multi modal Large Language Model}
Multi-modal Large Models: With the advancement of large models, many lightweight large models such as LLAMA \cite{touvron2023llama} and Mistral \cite{jiang2023mistral} have provided possibilities for usage and deployment in low-resource scenarios. The combination of large language models and vision encoders enables these models to effectively process tasks that involve multimodal data. Some studies \cite{li2023blip2,alayrac2022flamingo,dai2023instructblip,chen2024lion}, utilize visual mappers, such as basic linear layers, to map the features from vision encoders to token-level features of large language models with matching dimensions. Through this simple mapping technique, large models demonstrate exceptional performance in various multimodal tasks, such as image captioning and visual question answering. This suggests that multimodal large models can achieve excellent visual perception capabilities with only minor parameter fine-tuning. Fine-tuning large language models through instructions \cite{chen2024lion, dai2023instructblip} to become task-driven for various multimodal tasks is a widely adopted approach. By crafting specific instructions, large language models can generate relevant results tailored to the specific task type. Some studies\cite{jin2024promptmrg} have injected disease labels as prior information into radiology report generation frameworks; however, this approach limits the scalability of such frameworks

\subsubsection{Vision Language Pretraining}
Vision language models aim to align vision and language features through cross-modal interaction \cite{ALBEF, wang2022ofa, li2023blip2, CLIP}. One of the most influential studies is CLIP \cite{CLIP}, which utilizes contrastive learning to align paired image embeddings and text embeddings. CLIP achieves a significant performance improvement across several downstream tasks, including zero-shot classification and cross-modal retrieval. Some medical multimodal pretraining methods \cite{chexzero,medclip,gloria} have demonstrated strong performance on zero-shot disease classification tasks. Meanwhile, some studies\cite{chen2022align} have effectively enhanced the alignment and reasoning capabilities of multimodal pre-trained models by incorporating external medical knowledge. These approaches achieve inference for common diseases and the diagnosis of rare diseases without the need for extensive structured labels. This provides insights for our model to generate radiology reports.

\section{Methods}
The proposed TRRG consists of a two-stage training process. In the pretraining stage, we focus on disease-aware cross-modal fine-grained alignment between radiographs and corresponding radiology reports. Large language models typically facilitate the alignment between images and text through supervised signals at the token level. By utilizing sentence-level contrastive learning, our approach enhances the detailed capture of disease information by the vision encoder. Furthermore, in conjunction with our proposed clue injection module and cross-modal clue interaction module in the fine-tuning stage, our model demonstrates superior performance in both language generation and clinical effectiveness. At the same time, our proposed cross-modal clue interaction module effectively facilitates alignment between visual embeddings and disease clue embeddings.  The details of TRRG are shown in Fig. \ref{Figurestage}.
\begin{figure}[t]   
	\centering
	\includegraphics[width=\linewidth,scale=1.00]{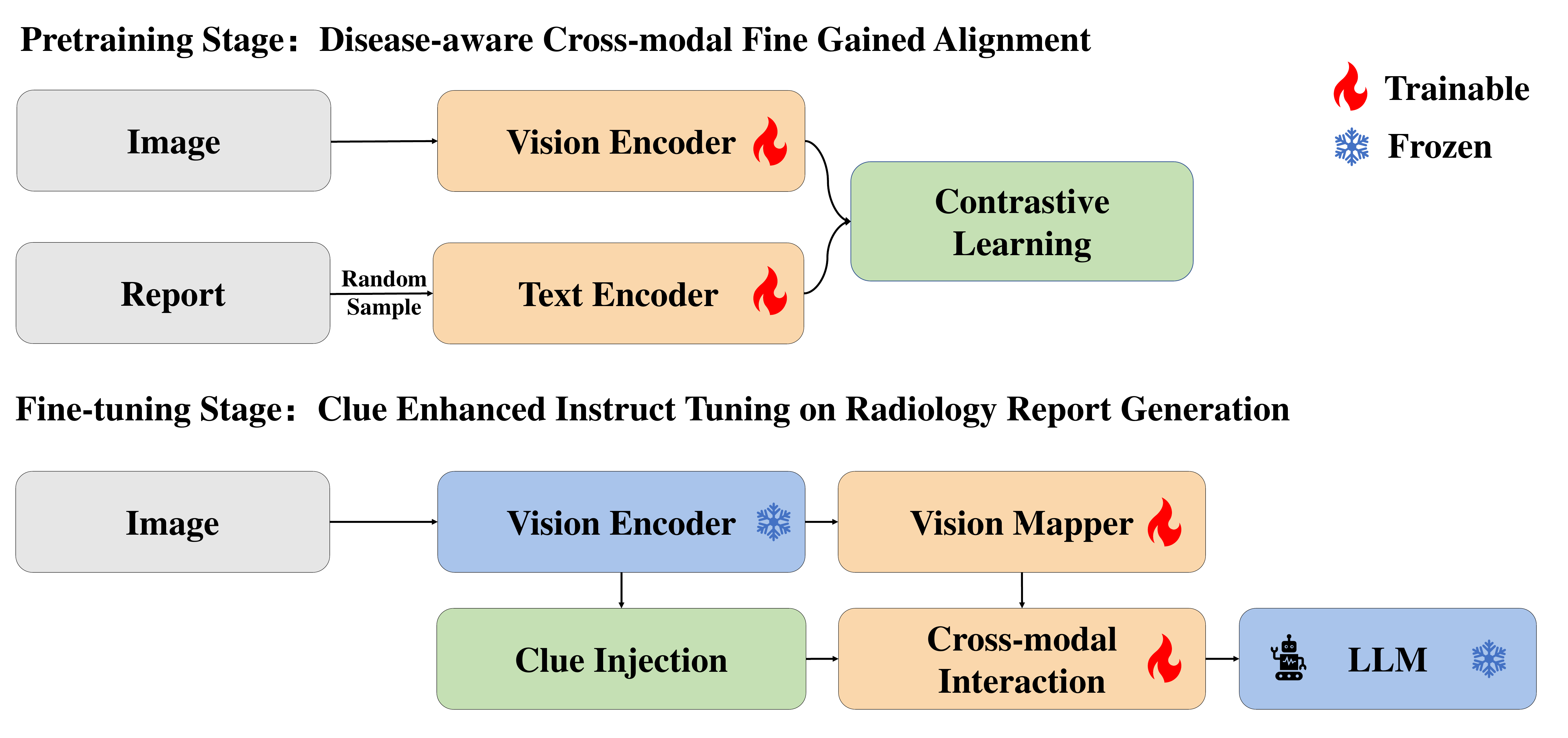}
	\caption{The training strategy of our proposed TRRG}
	\label{Figurestage}
 \vspace{-1em}
\end{figure}

\begin{figure*}[htbp]   
\vspace{-1em}
	\centering
	\includegraphics[scale=0.85]{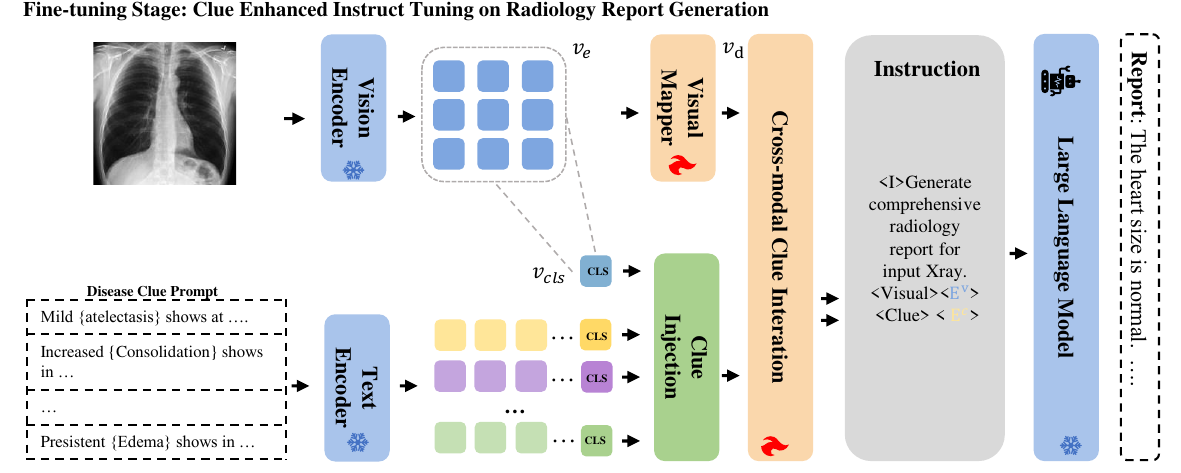}
	\vspace{-0.8em}
	\caption{During the fine-tuning stage, the visual encoder and the clue encoder are frozen, and disease clues are injected simultaneously through the clue injection module. In this stage, visual embeddings processed by the visual mapper interact with disease clue embeddings through cross-modal clue interaction. Finally, the frozen large language model is fine-tuned through instruction-based fine-tuning to achieve medical image report generation.}
	\label{mainfigure}
\vspace{-1em}
\end{figure*}

\subsection{Stage 1: Disease-aware Cross-modal Fine Gained Alignment}

Studies \cite{medclip, chexzero} demonstrate that pre-trained CLIP models, trained on large-scale medical image-text pairs, exhibit accuracy comparable to human performance in zero-shot disease classification tasks. Inspired by this, we decompose our pipeline into stage-wise training. During the pre-training stage, we utilize random sampling of sentences from radiology reports for training. This approach enhances the representation ability of vision encoders and also accommodates the robust clue prompting of disease clue injection modules. Given an Image-Text Pair $(I,T)$, where $T:\{T_1,T_2,...,T_t\}$, $T_i$ represents the sentence-level component of the radiology report. The images are encoded into $\mathbf{v}=\{\mathbf{v}_{cls},\mathbf{v}_{1},...,\mathbf{v}_{n}\} \in \mathbb{R}^{(n+1)*d}$ by the transformer-based vision encoder, where $n$ is the number of patches in the images and "cls" represents the pooling token of the outputs.
\begin{equation}
  \mathbf{v}=\mathbf{E}_{img}(I).   
\end{equation}

 For the sentence randomly sampled from the corresponding radiology report, a BERT-based model encodes it as $\mathbf{t}=\{\mathbf{t}_{cls},\mathbf{t}_{1},...,\mathbf{t}_{m}\} \in \mathbb{R}^{(m+1)*d}$, where $m$ is the maximum length of a sentence in the training corpus. We extract their global representations $\mathbf{t}\in \mathbb{R}^{1*d}$ by utilizing the "CLS" token of a BERT-like model.
\begin{equation}
\mathbf{t}=\mathbf{E}_{txt}(T_{r}),
\end{equation}
where $T_{r}$ is a randomly selected sentence from $T$. Finally, we compute the text-to-image contrastive loss and image-to-text contrastive loss to enhance our vision encoder's ability to learn better disease-oriented representations. We only utilize pooled visual tokens $\mathbf{v}_{cls}$ and textual tokens $\mathbf{v}_{cls}$. For an image embedding and text embedding $\{\mathbf{v}_i^{\prime},\mathbf{t}_i\}_{i=1}^N$, the optimizing objective is InfoNCE loss \cite{infoNCE}, which can be formulated as:
\begin{equation}
\mathcal{L}_{v2t}=-log(\frac{exp(\sigma(\mathbf{t}_i,\mathbf{v}_i^{\prime})/\tau)}{\sum_{j=1}^Nexp(\sigma(\mathbf{t}_i,\mathbf{v}_j^{\prime})/\tau)}),
\end{equation}

\begin{equation}
\mathcal{L}_{t2v}=-log(\frac{exp(\sigma(\mathbf{v}_i',\mathbf{t}_i)/\tau)}{\sum_{j=1}^Nexp(\sigma(\mathbf{v}_i',\mathbf{t}_j)/\tau)}).
\end{equation}

The total loss of the pretraining stage is calculated as follows, where $N$ represents the batch size, and $\tau$ is a temperature factor:
\begin{equation}
\mathcal{L}=\mathcal{L}_{v2t}+\mathcal{L}_{t2v}.
\end{equation}

\subsection{Stage 2: Clue Enhanced Instruct Tuning on Radiology Report Generation}

\subsubsection{Visual Embedding} 
Given an image $\mathbf{x} \in \mathbb{R}^{H \times W \times C}$, we reshape it into a sequence of flattened 2D patches $\mathbf{v} \in \mathbb{R}^{n \times d}$. The transformer-based frozen vision encoder encodes the input patch tokens into an encoded patch embedding $\mathbf{x}_e = \{\mathbf{v}_{cls}, \mathbf{v}_{1}, ..., \mathbf{v}_{n}\} \in \mathbb{R}^{(n+1) \times d}$. Subsequently, the patch embedding will be fed into a visual mapper layer composed of linear layers, which transform the patch embedding $\mathbf{v}_d \in \mathbb{R}^{n \times d}$ to the same dimension as the LLM's word embedding. Thus, we obtain generative vision embedding $\mathbf{v}_d \in \mathbb{R}^{n \times d}$ and visual disease expert tokens $\mathbf{v}_{cls} \in \mathbb{R}^{1 \times d}$ that have been encoded only by the frozen vision encoder. The visual embedding process can be expressed as follows:
\begin{equation}
\mathbf{v}_e=E_{img}(x),
\end{equation}
\begin{equation}
\mathbf{v}_d=W\mathbf{v}_e + b,
\end{equation}
where $W$ is the trainable weight of the visual mapper.
\begin{equation}
\mathbf{v}_{cls}= \frac{1}{n} \sum_{i=1}^{n} \mathbf{v}_i \quad \mathbf{v}_i \in \mathbb{R}^{1 \times d,
}
\end{equation}
$\mathbf{v}_i$ is an element of the path token sequences $\mathbf{v}_e$. After visual encoding, we obtain the disease visual embedding $\mathbf{v}_d \in \mathbb{R}^{n \times d}$ and the disease visual expert token $\mathbf{v}_{cls} \in \mathbb{R}^{1 \times d}$.

\subsubsection{Clue Injection Module}
 Following Gloria's work \cite{gloria}, we utilize a variety of language descriptions to construct disease clue prompts. By extracting multiple descriptive phrases from the dataset related to the severity and location of a specific disease, we randomly combined phrases associated with disease type, severity level, and location to create disease clue prompts. Specifically, our prompt construction template is as follows:
$$Clue: <severity> <disease> at <location>$$

We created manual templates for $m$ common diseases, such as opacity, pneumonia, and pneumothorax, to act as potential prompts for identifying diseases. Then, the frozen text encoder encoded disease clue prompts into multiple disease clue embeddings $\mathbf{c}:\{\mathbf{c}^i\}_{i=1}^m$, where $\mathbf{\mathbf{c}^i}=\{\mathbf{\mathbf{c}_{cls}},\mathbf{\mathbf{c}_1},....,\mathbf{\mathbf{c}_r}\}$, and $\mathbf{c}^i \in \mathbb{R}^{(r+1) \times d}$ is the i-th disease clue embedding. Here, $m$ represents the number of diseases defined, and $r$ is the maximum length of disease clue prompts. Subsequently, we compute clue weights, which represent the importance of disease clue expert token $\mathbf{c}_{cls}^i\in\mathbb{R}^{1*d}$ to visual disease expert tokens $\mathbf{v}_{cls}\in\mathbb{R}^{1*d}$ through matrix multiplication. The computing process of clue weight can be formulated as follows:

\begin{equation}
w_{i}=\text{softmax} \left( \mathbf{v}_{cls} \cdot \mathbf{c}_{cls}^i \right) \quad i\in 1,.....m ,
\end{equation}
where clue weight $w: \{w_1, w_2, \cdots, w_m\} \in \mathbb{R}^{1 \times m}$, where $m$ represents the number of different disease clues. We assign weights to each clue to represent the importance of this disease relative to the images. After that, we utilize the Hadamard product to obtain the weighted disease clue embedding. The weighted clue embeddings are computed as follows:
\begin{equation}
\mathbf{c}^i=w_i \odot \mathbf{c}^i .
\end{equation}

Due to some clues being irrelevant to the corresponding images, we only select the top-k important clues as disease expert clues for the final model input. Our final injection clues are:
\begin{equation}
\mathbf{c}_s: \{\mathbf{c}^{i}\} \quad i \in \mathbf{topk}(w, k),
\end{equation}
where $\mathbf{topk}(w, k)$ presents the indices of the top-k highest values in the set $w$. Finally, we consider the multiple disease clue embedding $\mathbf{c}_s \in {\mathbb{R}^{k\times r \times d}}$ as our injection clues.
\begin{figure}[t]   
	\centering
	\includegraphics[width=1\linewidth,scale=1.00]{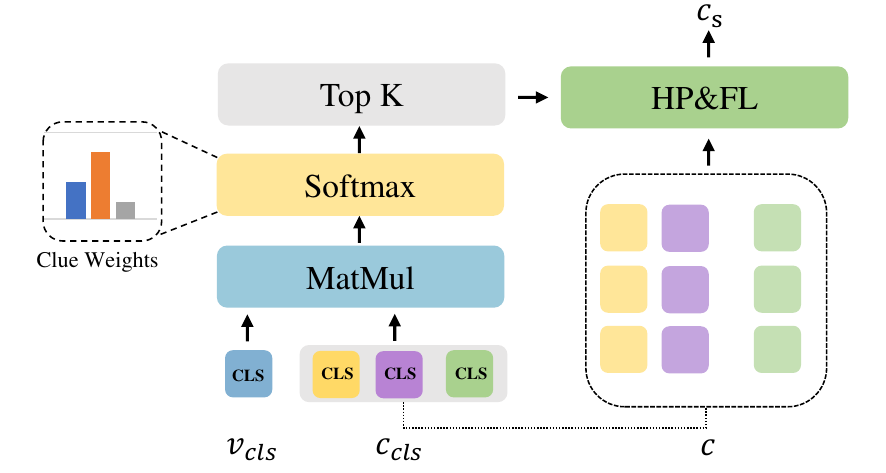}
	\caption{Architecture of Clue Injection Module, HP and FL represent Hadamard Product and Flatten, respectively.}
	\label{Figurestage}
 \vspace{-1em}
\end{figure}

\subsubsection{Cross Modal Clue Interaction}
For multi-disease clues $\mathbf{c}_s \in \mathbb{R}^{k\times r \times d}$, these clues are highly aligned with visual representations obtained through the frozen vision encoder but do not fully interact with the generated visual features obtained from the visual mapper. Therefore, we propose a Cross-Modal Clue Interaction Module to simultaneously enhance generative representations and facilitate cross-modal interaction of disease clues. Typically, there is often a significantly larger number of disease clue tokens compared to visual tokens. To address the excessive disease clue token input, we define a set of learnable queries for cross-modal interaction.

Finally, we propose a Disease Clue Consistency Loss to maintain sufficient attention to disease clue embeddings in cross-modal interaction and provide disease-oriented supervision signals for fine-tuning with large language models. Attention mechanisms are adopted in the multimodal interaction module. We adopt a two-stream architecture for cross-modal feature interaction. Each structure comprises a self-attention module, a cross-attention module, and a Feed Forward layer. The self-attention layer facilitates intra-modal interaction, enhancing feature representation within each modality. Conversely, the cross-attention mechanism ensures alignment between textual clues and visual representations, facilitating cross-modal interaction by enforcing consistency across different modalities. The attention is defined as attention:
\begin{equation}
    \operatorname{Attn}(Q,K,V)=\operatorname{softmax}(\frac{QK^{\top}}{\sqrt{d_k}})V .
\end{equation}

For visual embedding, $\mathbf{v}_{d}=\{\mathbf{v}_{1},...,\mathbf{v}_{n}\}$, disease clue embedding $\mathbf{c}_s \in {\mathbb{R}^{k\times r \times d}}$, We flatten the disease clues into  clue tokens $\mathbf{c}_s=\{\mathbf{c}_{1},...,\mathbf{c}_{(k\times r)}\}$, where $\mathbf{c}_i \in \mathbb{R}^{1 \times d}$.
Next, both visual embeddings and clue embeddings are fed into linear projection layer:
\begin{equation}
Q^{v}=W_{Q}^{v}\mathbf{v}_{d},K^{v}=W_{K}^{v}\mathbf{v}_{d},V^{v}=W_{V}^{v}\mathbf{v}_{d}  ,  
\end{equation}
\begin{equation}
Q^{c}=W_{Q}^{c}\mathbf{c}_{s},K^{c}=W_{K}^{c}\mathbf{c}_{s},V^{c}=W_{V}^{c}\mathbf{c}_{s} ,   
\end{equation}
\begin{equation}
\mathbf{V}^{\prime}=\operatorname{Attn}(Q^{v},K^{v},V^{v}),
\end{equation}
\begin{equation}
\mathbf{C}^{\prime}=\operatorname{Attn}(Q^{c},K^{c},V^{c}),   
\end{equation}
where $\mathbf{V}^{\prime},\mathbf{C}^{\prime}$ are visual embeddings and disease clue embeddingss after multi-head self attention layer. Next, we utilize learnable tokens $E=\{E_1, E_2, ..., E_L\}$ as a common feature space to establish associations between the visual and textual modalities, where $L$ represents the number of learnable tokens. In detail, we employ a scaled dot-product attention layer to calculate the correlation between the learnable tokens $E$ and the mapped visual tokens $E^v$. We perform the same operation on learnable queries and disease clue embeddings and obtained clue tokens ${E}^c$. This process can be expressed as:
\begin{equation}
\mathrm{E^{e}}=\mathrm{Attn}(E,E,E) ,
\end{equation}
\begin{equation}
\mathrm{E}^v=\mathrm{FFN(Attn}(E^{e},\mathbf{V}^{\prime},\mathbf{V}^{\prime})),
\end{equation}
\begin{equation}
\mathrm{E}^c=\mathrm{FFN(Attn}(E^{e},\mathbf{C}^{\prime},\mathbf{C}^{\prime})).
\end{equation}
 Furthermore, since disease clues are often sparse during cross-modal interaction, to enhance the consistency between visual tokens and clue tokens and improve effective supervision signals during radiology report generation, we propose a disease-aware consistency loss. Our disease-aware consistency loss is calculated as:
\begin{equation}
    \mathcal{L}_{DC}=-\frac1K\sum_{i=1}^K\frac{E^v\cdot E^c }{\|E^v\|\|E^c\|},
\end{equation}

We calculate the similarity between visual tokens and clue tokens and aim to maximize the alignment between visual and textual tokens. The disease-aware consistency loss effectively endows visual tokens with the ability to perceive diseases.

\begin{figure}[t]   
	\centering
	\includegraphics[width=0.8\linewidth,scale=1.00]{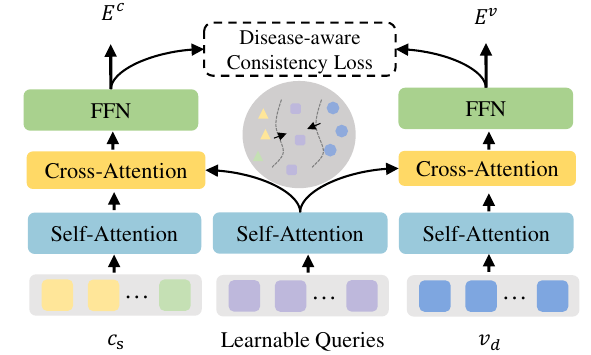}
   \vspace{-0.8em}
	\caption{Architecture of Cross Modal Clue Interaction Module}
	\label{Figurestage}
 \vspace{-1em}
\end{figure}
\subsubsection{Optimization Objective}
We train our model  by minimizing the negative log-likelihood of $\mathbf{P}(t)$ given the image features: 
\begin{equation}
    {\mathcal L}_{CE} = \sum_{i=1}^{T}log P_{\theta}({\mathbf t}_i|{E^v,E^c}, {t}_{i-1}, \cdots, {\mathbf t}_1) ,
\end{equation}
where based on the image embedding $E^v$ and clue embedding $E^c$  and the first $(i-1)$ words.
Our overall objective function is:
\begin{equation}
   {\mathcal L} = {\mathcal L}_{CE} +  {\mathcal L}_{DC} .
\end{equation}


\begin{table*}[t]

\vspace{-1em}
 \label{tab:mainresults}
 \caption{Comparisons of the proposed TRRG with previous studies on the \textsc{IU X-Ray} and MIMIC-CXR test set with respect to language generation (NLG) and clinical efficacy (CE) metrics. Best results are in \textbf{bold}.}	
  \vspace{-1.3em}
	\begin{center}
		\resizebox{0.98\textwidth}{!}{\begin{tabular}{c|c|ccccccc|ccc}
			\hline
			\multirow{2}{*}{Dataset} & \multirow{2}{*}{Model} & \multicolumn{7}{c|}{NLG Metrics} & \multicolumn{3}{c}{CE Metrics}\\
			& & BLEU-1 & BLEU-2 & BLEU-3 & BLEU-4& ROUGE & METEOR& CIDEr  & Precision & Recall & F1 \\
            
			\hline
			\multirow{12}{*}{\textsc{IU X-Ray}} 
          &  HGRG-Agent \cite{li2018hybrid}    &  0.438          &  0.298          &  0.208          &  0.151          &  0.322  &  -              &  0.343      & - & - & -     \\
          &  KERP \cite{Knowledge-Driven}          &  0.482          &  0.325        &  0.226          &  0.162          &  0.339  &  -              &  0.280      & - & - & -   \\
          & R2Gen~\cite{chen2020generating}          & 0.470          & 0.304          & 0.219          & 0.165          & 0.371  & 0.187       & -   &- &- &-              \\

          &  PPKED~\cite{CVPR201_PPKD}          &  0.483          &  0.315          &  0.224          &  0.168         &  0.376  &  0.187          &  0.351      & - & - & -   \\
          &  GSK~\cite{2021Knowledge}         &  \textbf{0.496}           &  \textbf{0.327}          &  \textbf{0.238}          &  \textbf{0.178}          & \textbf{0.381}  & -        &  0.382           & - & - & - \\
           & R2GenCMN~\cite{chen2022cross}          & 0.475          & 0.309          & 0.222          & 0.170          & 0.375  & 0.191              & -          & - & - & -\\
        & METransformer~\cite{wang2023metransformer}        & 0.483          & 0.322          & 0.228          & 0.172          & 0.380  & 0.192          & \textbf{0.435}
          & - & - & - \\
          \cline{2-12} 
        & \textbf{TRGG (Ours)}         & 0.482         & 0.302        & 0.217          & 0.151          & 0.377  & \textbf{0.209}               &0.405         & - & - & - \\ 
			\hline
			\multirow{12}{*}{MIMIC-CXR} 
          & M2Transformer~\cite{2020Meshed} & 0.332              & 0.210              & 0.142              & 0.101              & 0.264      & 0.134              & 0.142              & - & - & -\\
          & R2Gen~\cite{chen2020generating}         & 0.353          & 0.218          & 0.145          & 0.103          & 0.277  & 0.142          & -         & 0.333 & 0.273 & 0.276 \\
           & PPKED~\cite{CVPR201_PPKD}         & 0.36           & 0.224          & 0.149          & 0.106          & 0.284  & 0.149          & 0.237             & - & - & - \\

          & GSK~\cite{2021Knowledge}         & 0.363           & 0.228          & 0.156          & 0.115          & 0.284  & -          & 0.203             & - & - & - \\
 & R2GenCMN~\cite{chen2022cross}         & 0.353          & 0.218          & 0.148          & 0.106          & 0.278  & 0.142          & -         & 0.334 & 0.275 & 0.278 \\
         
          & MSAT~\cite{MSAT}        & 0.373           & 0.235          & 0.162          & 0.120         & 0.282  & 0.143          & 0.299
          & - & - & - \\
           & METransformer~\cite{wang2023metransformer}        & 0.386           & 0.250          & 0.169          & 0.124          & 0.291  & 0.152          & \textbf{0.362}
          & 0.364 & 0.309 & 0.311 \\
                     & DCL~\cite{li2023dynamic}        & - 
           & -         & -         & 0.107          & 0.284  & 0.150          & 0.281
          & \textbf{0.471 }&0.352 &0.373 \\
           & R2GenGPT~\cite{R2GenGPT}        & 0.365 
                 
           & 0.237          & 0.163          & 0.117          & 0.277  & 0.136          & 0.145
          & 0.341 &0.312 &0.325 \\
          & FGIRG~\cite{chen2024fine}   
          & 0.379           & 0.234          & 0.154          & 0.106          &  0.285  &0.162        & -
          &- &-& - \\
          & R2GMMN~\cite{R2GMMN}   
          &   0.396         & 0.244           & 0.162          & 0.115         &  0.274   & 0.151        & -
          &0.411&0.398& 0.389 \\

          \cline{2-12} 
          & \textbf{TRGG (Ours)}          & \textbf{0.436} & \textbf{0.298} & \textbf{0.213} & \textbf{0.157} & \textbf{0.336}  & \textbf{0.167} & 0.219 & 0.403 & \textbf{0.399} & \textbf{0.393}\\
			\hline
		\end{tabular}}
	\end{center}

\vspace{-2em}
\end{table*}

\section{Experiment}
\subsection{Datasets and Evaluation Metrics}
\subsubsection{Datasets}
\textbf{IU-Xray} \cite{2015iu-xray} is a widely recognized benchmark dataset for radiology report generation. The dataset consists of over 7,470 chest X-ray images and 3,955 corresponding radiology reports manually annotated by expert radiologists. \textbf{MIMIC-CXR} \cite{2019MIMIC} is a dataset comprising 64,588 patients collected at the Beth Israel Deaconess Radiology Center between 2011 and 2016. It includes 77,110 chest X-ray images and 227,835 corresponding free-text radiology reports. To ensure experimental fairness, we replicated the experimental settings of previous studies. This led to a training set of 222,758 samples, with validation and test sets comprising 1,808 and 3,269 samples, respectively.

\subsubsection{Evaluation Metrics}
Based on previous research, we evaluate our proposed radiology report generation model from two perspectives.
1) Evaluation of Language Generation Quality (\textbf{NLG Metrics}): Utilizing commonly used linguistic evaluation metrics such as BLEU \cite{papineni2002bleu}, Rouge-L \cite{lin-2004-rouge}, and CIDEr \cite{vedantam2015cider}. 2) Clinical Effectiveness Metrics (\textbf{CE Metrics}): We employ NLP text disease labeler ChexBERT\cite{smit2020chexbert} for text classification. We extract 14 common diseases from the generated reports and reference report. Precision, recall, and F1 score are used to assess performance in terms of clinical efficacy.

\section{Main Results}

\subsubsection{Implementation Details}

We utilized the Mistral-7B \cite{jiang2023mistral} model as  large language model, Swin-Transformer \cite{liu2021swin} as the vision encoder, and ClinicalBERT \cite{alsentzer2019publicly} for the text encoder in the pretraining stage. For template construction, we utilized 14 common diseases and created templates for each of them. We set the maximum clue number, K to 3, and ultimately selected the top 3 diseases with the highest weights for disease clue injection. The dimensions of visual embedding and text embedding are both set to 1024. Additionally, in the cross-modal cue interaction module, the number of heads in the linear projection layer of the attention mechanism to be 8. Moreover, both the attention layer and the cross-attention layer were set to have one layer. The training process was conducted on four NVIDIA A40 48GB GPUs. We trained the model on the MIMIC-CXR dataset for 5 epochs and on the IU-Xray dataset for 20 epochs. The batch size was set to 8, and the learning rate was 1e-4.

We compare the performance of our model with a wide range of state-of-the-art models in image captioning and radiology report generation. Table \ref{tab:mainresults} presents the comparison results for both Natural Language Generation (NLG) metrics and CE metrics. The models we compare include R2Gen \cite{chen2020generating}, R2GenCMN \cite{chen2022cross}, PPKED \cite{CVPR201_PPKD}, R2GenGPT\cite{R2GenGPT}, FGIRG\cite{chen2024fine} and R2GMMN\cite{R2GMMN}.  It can be observed that our model outperforms the current state-of-the-art methods across various language generation metrics, especially on MIMIC-CXR. Our model's performance on the IU-Xray dataset is acceptable but not exceptional. This could be attributed to the limited size of the IU-Xray dataset, which consists of only 2.8K image-text pairs. Consequently, the scarcity of training data may hinder the effective learning of text generation capabilities during the fine-tuning of large language models.The larger sample size of the MIMIC-CXR dataset allows for more comprehensive training of the vision encoder during the pretraining stage, thereby facilitating more consistent cross-modal alignment. 

We achieved a significant improvement across all Natural Language Generation (NLG) metrics except for CIDER\cite{vedantam2015cider} . METransformer\cite{wang2023metransformer} employed a specialized optimization strategy involving a voting strategy for the CIDER metric, thereby achieving significant superiority in performance according to the CIDER metric. Our model acquire a significant ability to maintain language consistency and rich semantics when generating medical image reports. This suggests that our approach can accurately capture keywords in radiology reports. Further evaluation of clinical efficacy metrics demonstrates the significant potential of our proposed method. Compared to other methods that fine-tuning large language models for radiology report generation, such as R2GenGPT\cite{R2GenGPT}, our model achieves significant improvements in multiple clinical efficacy metrics without relying on any external disease annotations. We observe that our proposed method achieves an average accuracy, recall, and F1 score of 0.403, 0.399, and 0.393, respectively, across various diseases. This represents a significant improvement over the current state-of-the-art method R2GMMN\cite{R2GMMN}, demonstrating the effectiveness of our approach. Furthermore, it indicates that our proposed model has stronger disease perception capabilities, enabling the generation of more truthful radiology reports.
\begin{figure*}[ht]  
	\centering
    
	\includegraphics[width=1\textwidth]{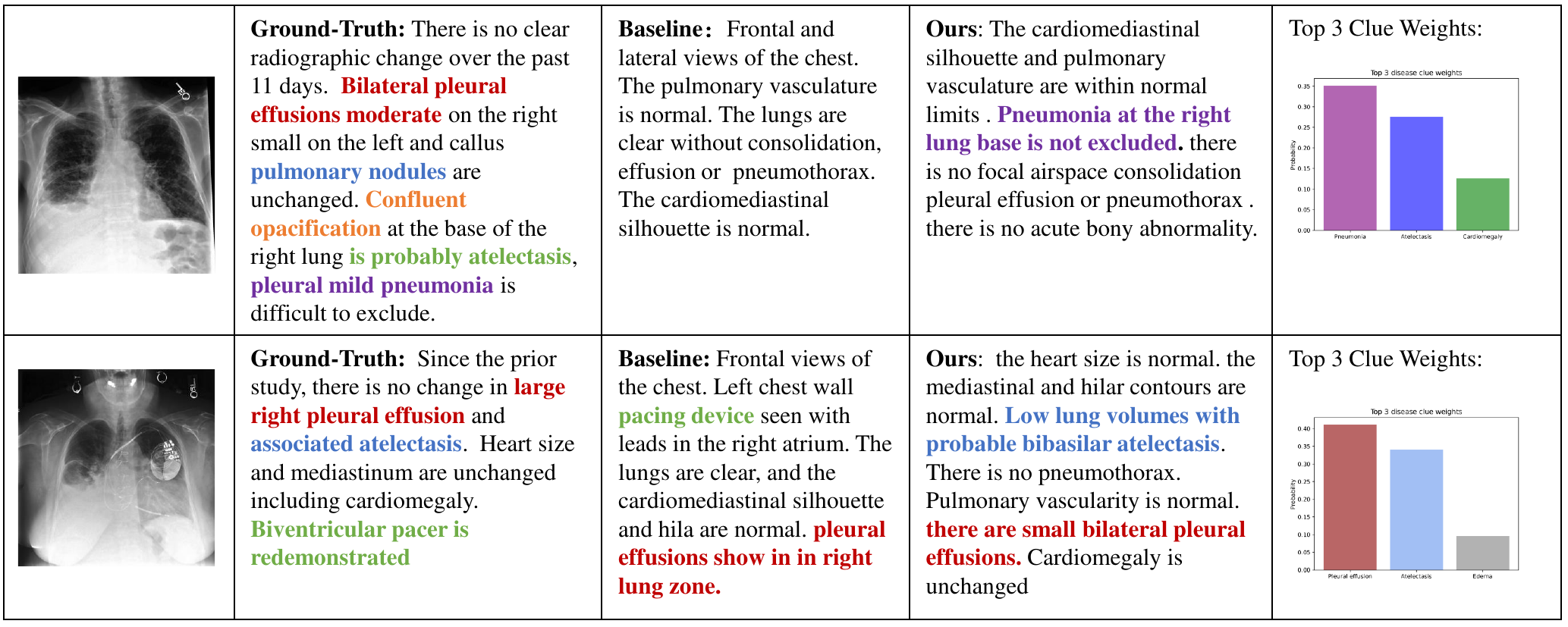}
   \vspace{-2em}
	\caption{We compare the generated results of the base model and the TRRG (Ours) with the ground truth, highlighting key information using colored fonts. Our model effectively generated specific descriptions tailored to diseases.}

	\label{Figure4}
\end{figure*}

\subsection{Ablation Study}

\textbf{Effectiveness of each component.}
We constructed our baseline model, called the "BASE" model, by solely fine-tuning the visual mapper during training using visual features. Meanwhile, we introduced a disease clue injection module, a cross-modal clue interaction module, and a disease-aware consistency loss function, abbreviated as "DCI," "CMCI" and "DAL" respectively. The symbol "+" denotes the experimental effects of adding different components to the base model, and specific results are presented in Tab. \ref{tab:ablation_study}. It can be observed that the three components we proposed all have a positive impact on performance. Despite some randomness in the experiments conducted in the deep learning laboratory, the results demonstrate credible comparability due to the systematic training partition applied to the MIMIC-CXR \cite{2019MIMIC} dataset. Our proposed three modules have achieved significant improvements of 14.5\%, 15.8\%, and 13.6\%, 11.7\%, 9.9\% in BLEU-4, ROUGE-L, METEOR, CIDEr, and F1 from the "Base" to our TRRG. The incorporation of different components resulted in varying degrees of improvement on the base model, thus validating the effectiveness of our approach.
\begin{table*}[t]
\caption{Ablation study of different componet we proposed, "DCI," "CMCI," and "DAL" respectively denote the Disease Clue Injection module, Cross-Modal Clue Interaction module, and Disease-Aware Loss function.}~\label{tab:ablation_study}
\vspace{-1em}
\centering
\scalebox{0.95}{
\begin{tabular}{ll|ccccc|ccccc}
\hline
\multicolumn{1}{c}{\multirow{2}{*}{\#}} & \multicolumn{1}{c}{\multirow{2}{*}{Models}} & \multicolumn{5}{c}{IU-Xray}                      & \multicolumn{5}{c}{MIMIC-CXR}                    \\ \cline{3-12} 
\multicolumn{1}{c}{}        & \multicolumn{1}{c}{}                        & BLEU-4 & ROUGE & METEOR & CIDEr&F1  & BLEU\_4 & ROUGE & METEOR & CIDEr&F1  \\ \hline
1                                       & \multicolumn{1}{l|}{BASE}               & \textbf{0.156}   & 0.370 & 0.194  & 0.387       &-       & 0.137   & 0.290 & 0.147  & 0.196  &0.354            \\
2                                       & \multicolumn{1}{l|}{+DCI}                    & 0.152   & 0.365 & 0.197  & 0.390 &-          & 0.142   & 0.311 & 0.156  & 0.207   &0.384    \\
3                                       & \multicolumn{1}{l|}{+DCI+CMCI}                & 0.147   & 0.372 & 0.205  & 0.402 &-         & \textbf{0.159}   & 0.324 & 0.162  & 0.211      & 0.387   \\
4                       & \multicolumn{1}{l|}{+DCI+CMCI+DAL}         & 0.151   & \textbf{0.377} & \textbf{0.209}  & \textbf{0.405}    &-    & 0.157   & \textbf{0.336} & \textbf{0.167}  & \textbf{0.219}   &\textbf{0.393}    \\ \hline
\end{tabular}}
\vspace{-1em}
\end{table*}

\textbf{Impact of clue numbers.}
The number of disease clue injections determines the richness of disease information perceived by the model during the fine-tuning stage. We set the value of \( k \) to 1, 2, 3, 4, and 5, and verified the corresponding experimental results on MIMIC-CXR \cite{2019MIMIC}. The experiments demonstrate that when our parameter \( k \) set to 3,  the model achieves superior performance across multiple evaluation metrics. This aligns with our intuitive understanding that, typically, a chest X-ray may exhibit concurrent manifestations of 3-4 diseases at most. As \( k \) gradually increases from 1 to 3, the model's performance steadily improves. When the parameter k exceeds 3, there is a decline in model performance. This phenomenon may be attributed to an excessive injection of irrelevant disease cues, which leads the model to inadequately attend to crucial regions within the images and pertinent disease cues.
\begin{table}[t]
\caption{Model with different disease clue numbers $k$, all experiments were conducted concerning varying quantities of disease clue injections on MIMI-CXR.}~\label{tab:ce_metrics}
\centering
\vspace{-0.3em}
\setlength{\tabcolsep}{2mm}{
\begin{tabular}{c|ccccc}
\hline
$k$ & BLEU-4 & ROUGE & METEOR &F1    \\ \hline
1   & 0.152     & 0.312  & 0.157 4&0.369 \\
2      & 0.155     &\textbf{0.341}  & 0.162 &0.377\\

\textbf{3}  & \textbf{0.157}   & 0.336 & \textbf{0.167}  &\textbf{0.393} \\
4       & 0.153     & 0.327  & 0.159&0.390\\
5    & 0.149     & 0.315  & 0.156 &0.392\\ \hline

\end{tabular}}
\end{table}

\textbf{Impact of the length of learnable queries.}
We conducted an ablation study on the length of learnable queries to explore the impact of different levels of compression of disease cues on the final model performance. We set the length  $L$ to 4, 8, 16, and 32, respectively, and found that the model achieved the best performance when the length of the learnable queries was set to 16. The appropriate length of learnable queries can ensure effective detection and compression of disease cues by the model.

\begin{table}[t]
\caption{Model with different learnable queries numbers $L$.}~\label{tab:proto}
\vspace{-1.8em}
\centering
\setlength{\tabcolsep}{2mm}{
\begin{tabular}{c|ccccc}
\hline
$L$& BLEU-4 & ROUGE & METEOR &F1    \\ \hline
4   & \textbf{0.161}     & 0.322  & 0.161&0.358 \\
8      & 0.155     &0.332  & 0.162 &0.377\\

\textbf{16}    & 0.157   &\textbf{0.336} & \textbf{0.167}  &\textbf{0.393} \\
32    & 0.142     & 0.329  & 0.156 &0.386\\ \hline

\end{tabular}}
\end{table}

\subsection{Qualitative analysis}
We conduct a qualitative analysis to validate the effectiveness of the proposed model. As depicted in Fig \ref{Figure4}, our disease clues and probabilities are highlighted using colored fonts. Compared to the base model, our proposed model tends to include more disease-related content with injected clues during the process of generating radiology reports. This observation confirms that our model has higher clinical reliability. In the second example, the conventional model hinted at the presence of auxiliary devices in the image, but our model failed to provide relevant descriptions. This indicates some limitations of our proposed approach, which may lead to an overemphasis on disease-related content in certain cases, thereby compromising findings and obscuring the expression of basic descriptions.
\vspace{-1em}

\section{Conclusion}
\vspace{-0.3em}
In this paper, we propose the TRGG for truthful radiology report generation based on fine-tuning large language models with injected disease cues. Our proposed stage-wise training strategy effectively promotes cross-modal alignment between radiography and reports. The clue injection module and cross-modal clue interaction module proposed by us can effectively facilitate the semantic representation of diseases and cross-modal alignment. Experimental results demonstrate the superiority of our approach. Extensive ablation studies and qualitative analyses confirm the effectiveness of our proposed method. Future research directions include developing a generalizable method for medical image report generation that can be applied across various medical imaging text report datasets, enabling further extension to heterogeneous modalities such as CT, MRI, and Ultrasound.
\bigskip

\bibliography{aaai25}

\begin{thebibliography}{53}
\providecommand{\natexlab}[1]{#1}

\bibitem[{Alayrac et~al.(2022)Alayrac, Donahue, Luc, Miech, Barr, Hasson, Lenc, Mensch, Millican, Reynolds et~al.}]{alayrac2022flamingo}
Alayrac, J.-B.; Donahue, J.; Luc, P.; Miech, A.; Barr, I.; Hasson, Y.; Lenc, K.; Mensch, A.; Millican, K.; Reynolds, M.; et~al. 2022.
\newblock Flamingo: a visual language model for few-shot learning.
\newblock In \emph{NeurIPS}.

\bibitem[{Alsentzer et~al.(2019)Alsentzer, Murphy, Boag, Weng, Jin, Naumann, and McDermott}]{alsentzer2019publicly}
Alsentzer, E.; Murphy, J.~R.; Boag, W.; Weng, W.-H.; Jin, D.; Naumann, T.; and McDermott, M. 2019.
\newblock Publicly available clinical BERT embeddings.
\newblock \emph{arXiv preprint arXiv:1904.03323}.

\bibitem[{Chen et~al.(2024{\natexlab{a}})Chen, Shen, Shao, Deng, and Nie}]{chen2024lion}
Chen, G.; Shen, L.; Shao, R.; Deng, X.; and Nie, L. 2024{\natexlab{a}}.
\newblock LION: Empowering Multimodal Large Language Model with Dual-Level Visual Knowledge.
\newblock In \emph{IEEE Conference on Computer Vision and Pattern Recognition (CVPR)}.

\bibitem[{Chen et~al.(2024{\natexlab{b}})Chen, Shen, Lin, Luo, Li, and Yuan}]{chen2024fine}
Chen, W.; Shen, L.; Lin, J.; Luo, J.; Li, X.; and Yuan, Y. 2024{\natexlab{b}}.
\newblock Fine-Grained Image-Text Alignment in Medical Imaging Enables Explainable Cyclic Image-Report Generation.
\newblock In \emph{62nd Annual Meeting of the Association for Computational Linguistics (ACL 2024)}.

\bibitem[{Chen, Li, and Wan(2022)}]{chen2022align}
Chen, Z.; Li, G.; and Wan, X. 2022.
\newblock Align, reason and learn: Enhancing medical vision-and-language pre-training with knowledge.
\newblock In \emph{Proceedings of the 30th ACM International Conference on Multimedia}, 5152--5161.

\bibitem[{Chen et~al.(2022)Chen, Shen, Song, and Wan}]{chen2022cross}
Chen, Z.; Shen, Y.; Song, Y.; and Wan, X. 2022.
\newblock Cross-modal memory networks for radiology report generation.

\bibitem[{Chen et~al.(2020)Chen, Song, Chang, and Wan}]{chen2020generating}
Chen, Z.; Song, Y.; Chang, T.-H.; and Wan, X. 2020.
\newblock Generating radiology reports via memory-driven transformer.
\newblock In \emph{EMNLP}.

\bibitem[{Cornia et~al.(2020{\natexlab{a}})Cornia, Stefanini, Baraldi, and Cucchiara}]{cornia2020meshedmemory}
Cornia, M.; Stefanini, M.; Baraldi, L.; and Cucchiara, R. 2020{\natexlab{a}}.
\newblock Meshed-Memory Transformer for Image Captioning.
\newblock In \emph{CVPR}.

\bibitem[{Cornia et~al.(2020{\natexlab{b}})Cornia, Stefanini, Baraldi, and Cucchiara}]{2020Meshed}
Cornia, M.; Stefanini, M.; Baraldi, L.; and Cucchiara, R. 2020{\natexlab{b}}.
\newblock Meshed-Memory Transformer for Image Captioning.
\newblock In \emph{CVPR}.

\bibitem[{Dai et~al.(2023)Dai, Li, Li, Tiong, Zhao, Wang, Li, Fung, and Hoi}]{dai2023instructblip}
Dai, W.; Li, J.; Li, D.; Tiong, A. M.~H.; Zhao, J.; Wang, W.; Li, B.; Fung, P.; and Hoi, S. 2023.
\newblock InstructBLIP: Towards General-purpose Vision-Language Models with Instruction Tuning.
\newblock In \emph{NeurIPS}.

\bibitem[{Dina et~al.(2015)Dina, Kohli, Rosenman, Shooshan, Laritza, Sameer, Thoma, and Mcdonald}]{2015iu-xray}
Dina, D.~F.; Kohli, M.~D.; Rosenman, M.~B.; Shooshan, S.~E.; Laritza, R.; Sameer, A.; Thoma, G.~R.; and Mcdonald, C.~J. 2015.
\newblock Preparing a collection of radiology examinations for distribution and retrieval.
\newblock \emph{Journal of the American Medical Informatics Association Jamia}.

\bibitem[{Huang et~al.(2021)Huang, Shen, Lungren, and Yeung}]{gloria}
Huang, S.-C.; Shen, L.; Lungren, M.~P.; and Yeung, S. 2021.
\newblock GLoRIA: A Multimodal Global-Local Representation Learning Framework for Label-efficient Medical Image Recognition.
\newblock In \emph{2021 IEEE/CVF International Conference on Computer Vision (ICCV)}.

\bibitem[{Jiang et~al.(2023)Jiang, Sablayrolles, Mensch, Bamford, Chaplot, Casas, Bressand, Lengyel, Lample, Saulnier et~al.}]{jiang2023mistral}
Jiang, A.~Q.; Sablayrolles, A.; Mensch, A.; Bamford, C.; Chaplot, D.~S.; Casas, D. d.~l.; Bressand, F.; Lengyel, G.; Lample, G.; Saulnier, L.; et~al. 2023.
\newblock Mistral 7B.
\newblock \emph{arXiv preprint arXiv:2310.06825}.

\bibitem[{Jin et~al.(2024)Jin, Che, Lin, and Chen}]{jin2024promptmrg}
Jin, H.; Che, H.; Lin, Y.; and Chen, H. 2024.
\newblock Promptmrg: Diagnosis-driven prompts for medical report generation.
\newblock In \emph{Proceedings of the AAAI Conference on Artificial Intelligence}, volume~38, 2607--2615.

\bibitem[{Johnson et~al.(2019)Johnson, Pollard, Greenbaum, Lungren, Deng, Peng, Lu, Mark, Berkowitz, and Horng}]{2019MIMIC}
Johnson, A. E.~W.; Pollard, T.~J.; Greenbaum, N.~R.; Lungren, M.~P.; Deng, C.~Y.; Peng, Y.; Lu, Z.; Mark, R.~G.; Berkowitz, S.~J.; and Horng, S. 2019.
\newblock MIMIC-CXR: A large publicly available database of labeled chest radiographs.
\newblock \emph{CoRR}.

\bibitem[{Johnson, Karpathy, and Fei{-}Fei(2016)}]{johnson2015densecap}
Johnson, J.; Karpathy, A.; and Fei{-}Fei, L. 2016.
\newblock DenseCap: Fully Convolutional Localization Networks for Dense Captioning.
\newblock In \emph{CVPR}.

\bibitem[{Li et~al.(2019{\natexlab{a}})Li, Liang, Hu, and Xing}]{li2019knowledgedriven}
Li, C.~Y.; Liang, X.; Hu, Z.; and Xing, E.~P. 2019{\natexlab{a}}.
\newblock Knowledge-driven Encode, Retrieve, Paraphrase for Medical Image Report Generation.

\bibitem[{Li et~al.(2019{\natexlab{b}})Li, Liang, Hu, and Xing}]{Knowledge-Driven}
Li, C.~Y.; Liang, X.; Hu, Z.; and Xing, E.~P. 2019{\natexlab{b}}.
\newblock Knowledge-Driven Encode, Retrieve, Paraphrase for Medical Image Report Generation.
\newblock In \emph{AAAI}.

\bibitem[{Li et~al.(2023{\natexlab{a}})Li, Li, Savarese, and Hoi}]{li2023blip2}
Li, J.; Li, D.; Savarese, S.; and Hoi, S. 2023{\natexlab{a}}.
\newblock Blip-2: Bootstrapping language-image pre-training with frozen image encoders and large language models.
\newblock In \emph{ICML}.

\bibitem[{Li et~al.(2021)Li, Selvaraju, Gotmare, Joty, Xiong, and Hoi}]{ALBEF}
Li, J.; Selvaraju, R.; Gotmare, A.; Joty, S.; Xiong, C.; and Hoi, S. C.~H. 2021.
\newblock Align before fuse: Vision and language representation learning with momentum distillation.
\newblock \emph{Advances in neural information processing systems}, 34: 9694--9705.

\bibitem[{Li et~al.(2023{\natexlab{b}})Li, Lin, Chen, Lin, Liang, and Chang}]{li2023dynamic}
Li, M.; Lin, B.; Chen, Z.; Lin, H.; Liang, X.; and Chang, X. 2023{\natexlab{b}}.
\newblock Dynamic Graph Enhanced Contrastive Learning for Chest X-ray Report Generation.
\newblock In \emph{Proceedings of the IEEE/CVF Conference on Computer Vision and Pattern Recognition}, 3334--3343.

\bibitem[{Li et~al.(2018)Li, Liang, Hu, and Xing}]{li2018hybrid}
Li, Y.; Liang, X.; Hu, Z.; and Xing, E.~P. 2018.
\newblock Hybrid Retrieval-Generation Reinforced Agent for Medical Image Report Generation.
\newblock In \emph{NeurIPS}.

\bibitem[{Lin(2004)}]{lin-2004-rouge}
Lin, C.-Y. 2004.
\newblock {ROUGE}: A Package for Automatic Evaluation of Summaries.
\newblock In \emph{ACL}.

\bibitem[{Liu et~al.(2021{\natexlab{a}})Liu, Wu, Ge, Fan, and Zou}]{CVPR201_PPKD}
Liu, F.; Wu, X.; Ge, S.; Fan, W.; and Zou, Y. 2021{\natexlab{a}}.
\newblock Exploring and Distilling Posterior and Prior Knowledge for Radiology Report Generation.
\newblock In \emph{CVPR}.

\bibitem[{Liu et~al.(2021{\natexlab{b}})Liu, Lin, Cao, Hu, Wei, Zhang, Lin, and Guo}]{liu2021swin}
Liu, Z.; Lin, Y.; Cao, Y.; Hu, H.; Wei, Y.; Zhang, Z.; Lin, S.; and Guo, B. 2021{\natexlab{b}}.
\newblock Swin transformer: Hierarchical vision transformer using shifted windows.
\newblock In \emph{Proceedings of the IEEE/CVF international conference on computer vision}, 10012--10022.

\bibitem[{Lu et~al.(2017)Lu, Xiong, Parikh, and Socher}]{2017Knowing}
Lu, J.; Xiong, C.; Parikh, D.; and Socher, R. 2017.
\newblock Knowing When to Look: Adaptive Attention via a Visual Sentinel for Image Captioning.
\newblock In \emph{CVPR}.

\bibitem[{Oord, Li, and Vinyals(2018)}]{infoNCE}
Oord, A.; Li, Y.; and Vinyals, O. 2018.
\newblock Representation Learning with Contrastive Predictive Coding.
\newblock \emph{Cornell University - arXiv,Cornell University - arXiv}.

\bibitem[{Pan et~al.(2020)Pan, Yao, Li, and Mei}]{pan2020xlinear}
Pan, Y.; Yao, T.; Li, Y.; and Mei, T. 2020.
\newblock X-Linear Attention Networks for Image Captioning.
\newblock In \emph{CVPR}.

\bibitem[{Papineni et~al.(2002)Papineni, Roukos, Ward, and Zhu}]{papineni2002bleu}
Papineni, K.; Roukos, S.; Ward, T.; and Zhu, W.-J. 2002.
\newblock Bleu: a method for automatic evaluation of machine translation.
\newblock In \emph{Proceedings of the 40th annual meeting of the Association for Computational Linguistics}, 311--318.

\bibitem[{Radford et~al.(2021)Radford, Kim, Hallacy, Ramesh, Goh, Agarwal, Sastry, Amanda, Mishkin, Clark, Krueger, and Sutskever}]{CLIP}
Radford, A.; Kim, J.; Hallacy, C.; Ramesh, A.; Goh, G.; Agarwal, S.; Sastry, G.; Amanda, A.; Mishkin, P.; Clark, J.; Krueger, G.; and Sutskever, I. 2021.
\newblock Learning Transferable Visual Models From Natural Language Supervision.
\newblock \emph{Cornell University - arXiv,Cornell University - arXiv}.

\bibitem[{Rennie et~al.(2017)Rennie, Marcheret, Mroueh, Ross, and Goel}]{rennie2017selfcritical}
Rennie, S.~J.; Marcheret, E.; Mroueh, Y.; Ross, J.; and Goel, V. 2017.
\newblock Self-Critical Sequence Training for Image Captioning.
\newblock In \emph{CVPR}.

\bibitem[{Shen et~al.(2024)Shen, Pei, Liu, and Tian}]{R2GMMN}
Shen, H.; Pei, M.; Liu, J.; and Tian, Z. 2024.
\newblock Automatic Radiology Reports Generation via Memory Alignment Network.
\newblock In \emph{Proceedings of the AAAI Conference on Artificial Intelligence}, volume~38, 4776--4783.

\bibitem[{Smit et~al.(2020)Smit, Jain, Rajpurkar, Pareek, Ng, and Lungren}]{smit2020chexbert}
Smit, A.; Jain, S.; Rajpurkar, P.; Pareek, A.; Ng, A.~Y.; and Lungren, M.~P. 2020.
\newblock CheXbert: Combining Automatic Labelers and Expert Annotations for Accurate Radiology Report Labeling Using BERT.
\newblock arXiv:2004.09167.

\bibitem[{Tang et~al.(2021)Tang, Wang, Liu, Rao, Li, and Li}]{tang2021clip4caption}
Tang, M.; Wang, Z.; Liu, Z.; Rao, F.; Li, D.; and Li, X. 2021.
\newblock Clip4caption: Clip for video caption.
\newblock In \emph{Proceedings of the 29th ACM International Conference on Multimedia}, 4858--4862.

\bibitem[{Tiu et~al.(2022)Tiu, Talius, Patel, Langlotz, Ng, and Rajpurkar}]{chexzero}
Tiu, E.; Talius, E.; Patel, P.; Langlotz, C.~P.; Ng, A.~Y.; and Rajpurkar, P. 2022.
\newblock Expert-level detection of pathologies from unannotated chest X-ray images via self-supervised learning.
\newblock \emph{Nature Biomedical Engineering}, 6(12): 1399--1406.

\bibitem[{Touvron et~al.(2023)Touvron, Lavril, Izacard, Martinet, Lachaux, Lacroix, Rozi{\`e}re, Goyal, Hambro, Azhar et~al.}]{touvron2023llama}
Touvron, H.; Lavril, T.; Izacard, G.; Martinet, X.; Lachaux, M.-A.; Lacroix, T.; Rozi{\`e}re, B.; Goyal, N.; Hambro, E.; Azhar, F.; et~al. 2023.
\newblock Llama: Open and efficient foundation language models.
\newblock \emph{arXiv preprint arXiv:2302.13971}.

\bibitem[{Vaswani et~al.(2017)Vaswani, Shazeer, Parmar, Uszkoreit, Jones, Gomez, Kaiser, and Polosukhin}]{vaswani2017attentionisallyouneed}
Vaswani, A.; Shazeer, N.; Parmar, N.; Uszkoreit, J.; Jones, L.; Gomez, A.~N.; Kaiser, L.; and Polosukhin, I. 2017.
\newblock Attention is All you Need.
\newblock In \emph{NIPS}.

\bibitem[{Vedantam, Zitnick, and Parikh(2015)}]{vedantam2015cider}
Vedantam, R.; Zitnick, C.~L.; and Parikh, D. 2015.
\newblock CIDEr: Consensus-based image description evaluation.
\newblock In \emph{CVPR}.

\bibitem[{Wang, Bhalerao, and He(2022)}]{wang2022cross}
Wang, J.; Bhalerao, A.; and He, Y. 2022.
\newblock Cross-modal prototype driven network for radiology report generation.
\newblock In \emph{European Conference on Computer Vision}, 563--579. Springer.

\bibitem[{Wang et~al.(2022{\natexlab{a}})Wang, Yang, Men, Lin, Bai, Li, Ma, Zhou, Zhou, and Yang}]{wang2022ofa}
Wang, P.; Yang, A.; Men, R.; Lin, J.; Bai, S.; Li, Z.; Ma, J.; Zhou, C.; Zhou, J.; and Yang, H. 2022{\natexlab{a}}.
\newblock Ofa: Unifying architectures, tasks, and modalities through a simple sequence-to-sequence learning framework.
\newblock In \emph{International Conference on Machine Learning}, 23318--23340. PMLR.

\bibitem[{Wang et~al.(2023{\natexlab{a}})Wang, Wang, Liu, Gao, Zhang, and Wang}]{wang2023self}
Wang, Y.; Wang, K.; Liu, X.; Gao, T.; Zhang, J.; and Wang, G. 2023{\natexlab{a}}.
\newblock Self adaptive global-local feature enhancement for radiology report generation.
\newblock In \emph{2023 IEEE International Conference on Image Processing (ICIP)}, 2275--2279. IEEE.

\bibitem[{Wang et~al.(2023{\natexlab{b}})Wang, Liu, Wang, and Zhou}]{wang2023metransformer}
Wang, Z.; Liu, L.; Wang, L.; and Zhou, L. 2023{\natexlab{b}}.
\newblock METransformer: Radiology Report Generation by Transformer with Multiple Learnable Expert Tokens.
\newblock In \emph{CVPR}, 11558--11567.

\bibitem[{Wang et~al.(2023{\natexlab{c}})Wang, Liu, Wang, and Zhou}]{R2GenGPT}
Wang, Z.; Liu, L.; Wang, L.; and Zhou, L. 2023{\natexlab{c}}.
\newblock R2gengpt: Radiology report generation with frozen llms.
\newblock \emph{Meta-Radiology}, 1(3): 100033.

\bibitem[{Wang et~al.(2022{\natexlab{b}})Wang, Tang, Wang, Li, and Zhou}]{MSAT}
Wang, Z.; Tang, M.; Wang, L.; Li, X.; and Zhou, L. 2022{\natexlab{b}}.
\newblock A medical semantic-assisted transformer for radiographic report generation.
\newblock In \emph{International Conference on Medical Image Computing and Computer-Assisted Intervention}, 655--664. Springer.

\bibitem[{Wang et~al.(2022{\natexlab{c}})Wang, Wu, Agarwal, and Sun}]{medclip}
Wang, Z.; Wu, Z.; Agarwal, D.; and Sun, J. 2022{\natexlab{c}}.
\newblock MedCLIP: Contrastive Learning from Unpaired Medical Images and Text.

\bibitem[{Wang et~al.(2021)Wang, Zhou, Wang, and Li}]{Wang_2021_CVPR}
Wang, Z.; Zhou, L.; Wang, L.; and Li, X. 2021.
\newblock A Self-Boosting Framework for Automated Radiographic Report Generation.
\newblock In \emph{CVPR}.

\bibitem[{Xue et~al.(2018)Xue, Xu, Long, Xue, Antani, Thoma, and Huang}]{2018Multimodal}
Xue, Y.; Xu, T.; Long, L.~R.; Xue, Z.; Antani, S.; Thoma, G.~R.; and Huang, X. 2018.
\newblock Multimodal Recurrent Model with Attention for Automated Radiology Report Generation.
\newblock In \emph{MICCAI}.

\bibitem[{Yang et~al.(2021)Yang, Wu, Ge, Zhou, and Xiao}]{2021Knowledge}
Yang, S.; Wu, X.; Ge, S.; Zhou, S.~K.; and Xiao, L. 2021.
\newblock Knowledge Matters: Radiology Report Generation with General and Specific Knowledge.
\newblock \emph{Medical Image Analysis}.

\bibitem[{Yang et~al.(2022)Yang, Wu, Ge, Zhou, and Xiao}]{KnowledgeMatters}
Yang, S.; Wu, X.; Ge, S.; Zhou, S.~K.; and Xiao, L. 2022.
\newblock Knowledge matters: Chest radiology report generation with general and specific knowledge.
\newblock \emph{Medical image analysis}, 80: 102510.

\bibitem[{Yin et~al.(2020)Yin, Qian, Wei, Li, Zhang, Li, and Zheng}]{2020icdmAutomatic}
Yin, C.; Qian, B.; Wei, J.; Li, X.; Zhang, X.; Li, Y.; and Zheng, Q. 2020.
\newblock Automatic Generation of Medical Imaging Diagnostic Report with Hierarchical Recurrent Neural Network.
\newblock In \emph{ICDM}.

\bibitem[{Yu et~al.(2019)Yu, Li, Yu, and Huang}]{2019MultimodalTransformer}
Yu, J.; Li, J.; Yu, Z.; and Huang, Q. 2019.
\newblock Multimodal transformer with multi-view visual representation for image captioning.
\newblock \emph{IEEE transactions on circuits and systems for video technology}, 30(12): 4467--4480.

\bibitem[{Yuan et~al.(2019)Yuan, Liao, Luo, and Luo}]{2019muiti-view}
Yuan, J.; Liao, H.; Luo, R.; and Luo, J. 2019.
\newblock Automatic Radiology Report Generation Based on Multi-view Image Fusion and Medical Concept Enrichment.
\newblock In \emph{MICCAI}.

\bibitem[{Zhang et~al.(2020)Zhang, Wang, Xu, Yu, Yuille, and Xu}]{2020When}
Zhang, Y.; Wang, X.; Xu, Z.; Yu, Q.; Yuille, A.; and Xu, D. 2020.
\newblock When radiology report generation meets knowledge graph.
\newblock In \emph{Proceedings of the AAAI conference on artificial intelligence}, volume~34, 12910--12917.

\end{thebibliography}

\newcommand{\answerYes}[1]{\textcolor{blue}{[#1]}} 
\newcommand{\answerNo}[1]{\textcolor{teal}{[#1]}} 
\newcommand{\answerNA}[1]{\textcolor{gray}{[#1]}} 
\newcommand{\answerTODO}[1]{\textcolor{red}{[#1]}}

\clearpage
\section{Reproducibility Checklist}

\answerYes{YES}
\answerNo{NO}
\answerNA{NA}

\begin{enumerate}

    \item This paper:
    \begin{enumerate}
        \item Includes a conceptual outline and/or pseudocode description of AI methods introduced. \answerYes{YES}
        \item Clearly delineates statements that are opinions, hypothesis, and speculation from objective facts and results. \answerYes{YES}
        \item Provides well marked pedagogical references for less-familiare readers to gain background necessary to replicate the paper. \answerYes{YES}
    \end{enumerate}

    \item Does this paper make theoretical contributions? (yes/no)
    \begin{enumerate}        
        \item All assumptions and restrictions are stated clearly and formally. \answerYes{YES}
        \item All novel claims are stated formally (\emph{e.g.}, in theorem statements). \answerYes{YES}
        \item Proofs of all novel claims are included. \answerYes{YES}
        \item Proof sketches or intuitions are given for complex and/or novel results. \answerYes{YES}
        \item Appropriate citations to theoretical tools used are given.\answerYes{YES}
        \item All theoretical claims are demonstrated empirically to hold. \answerYes{YES}
        \item All experimental code used to eliminate or disprove claims is included.\answerYes{YES}
    \end{enumerate}

    \item Does this paper rely on one or more datasets? \answerYes{YES}
    \begin{enumerate}
        \item A motivation is given for why the experiments are conducted on the selected datasets. \answerYes{YES}
        \item All novel datasets introduced in this paper are included in a data appendix. \answerNA{NA}
        \item All novel datasets introduced in this paper will be made publicly available upon publication of the paper with a license that allows free usage for research purposes.  \answerNA{NA}
        \item All datasets drawn from the existing literature (potentially including authors’ own previously published work) are accompanied by appropriate citations.\answerYes{YES}
        \item All datasets drawn from the existing literature (potentially including authors’ own previously published work) are publicly available. \answerYes{YES}
        \item All datasets that are not publicly available are described in detail, with explanation why publicly available alternatives are not scientifically satisficing. \answerNA{NA}
    \end{enumerate}

    \item Does this paper include computational experiments? (yes/no)
    \begin{enumerate}
        \item Any code required for pre-processing data is included in the appendix.  \answerNA{NA}
        \item All source code required for conducting and analyzing the experiments is included in a code appendix.  \answerYes{YES}
        \item All source code required for conducting and analyzing the experiments will be made publicly available upon publication of the paper with a license that allows free usage for research purposes. \answerYes{YES}
        \item All source code implementing new methods have comments detailing the implementation, with references to the paper where each step comes from. \answerYes{YES}
        \item If an algorithm depends on randomness, then the method used for setting seeds is described in a way sufficient to allow replication of results. \answerYes{YES}
        \item This paper specifies the computing infrastructure used for running experiments (hardware and software), including GPU/CPU models; amount of memory; operating system; names and versions of relevant software libraries and frameworks. \answerYes{YES}
        \item This paper formally describes evaluation metrics used and explains the motivation for choosing these metrics. \answerYes{YES}
        \item This paper states the number of algorithm runs used to compute each reported result. \answerNo{NO}
        \item Analysis of experiments goes beyond single-dimensional summaries of performance (e.g., average; median) to include measures of variation, confidence, or other distributional information. \answerNo{NO}
        \item The significance of any improvement or decrease in performance is judged using appropriate statistical tests (e.g., Wilcoxon signed-rank). \answerNo{NO}
        \item This paper lists all final (hyper-)parameters used for each model/algorithm in the paper’s experiments. \answerYes{YES}
        \item This paper states the number and range of values tried per (hyper-) parameter during development of the paper, along with the criterion used for selecting the final parameter setting. \answerNo{NO}
    \end{enumerate}
    
\end{enumerate}

\clearpage

\end{document}